\documentclass[10pt,twocolumn,letterpaper]{article}

\PassOptionsToPackage{table}{xcolor}
\usepackage[algorithms]{wacv} 
\usepackage{tabularx}

\usepackage{algorithm}
\usepackage{algpseudocode}

\usepackage{pgfplots}
\pgfplotsset{compat=1.18}

\definecolor{wacvblue}{rgb}{0.21,0.49,0.74}
\usepackage[pagebackref,breaklinks,colorlinks,allcolors=wacvblue]{hyperref}

\def\wacvPaperID{3460} 
\def\confName{WACV}
\def\confYear{2027}

\title{VisCurate: Evaluating Behavioral Equivalence of Visual Skills}

\author{Md Farhan Ishraq$^1$, Md Farhan Ishmam$^2$, Reaz Hassan Joarder$^1$, Md Hasanul Kabir$^1$\\
  $^1$Islamic University of Technology\quad $^2$University of Utah\\
  {\tt\small \{farhanishraq17,hasanul\}@iut-dhaka.edu}\\\vspace{2pt}
  \small \url{https://farhanishraq17.github.io/Viscurate-Website/}\vspace{-8pt}
}

\begin{document}
\maketitle
\begin{abstract}
Modern vision-based agents have shifted toward skill-based problem-solving, where a skill is an executable image-to-image operation, e.g., cropping, thresholding, or edge detection. As these skill libraries grow, they accumulate duplicated, mislabeled, and defective skills. The signals normally used to curate them (a skill's name, description, or source code) say little about how the skill visually changes the image. We present \textbf{VisCurate}, a framework and benchmark that curates visual skill libraries from observed execution rather than from text. VisCurate runs each candidate skill pair on a shared probe battery and assigns one of six behavioral relations, each tied to a library edit that is safe to make. The benchmark injects seven defect types into a clean library at controlled rates, producing 300 reproducible instances with ground-truth relations and repairs. While text baselines accept up to five of six false merges on hard negatives, our verifier accepts none and recovers 51.2\% of the pairs that are truly mergeable. On two real, uncurated libraries, VisCurate finds that 92\% of the skills a package shares with its downstream copy are redundant, while execution confirms only 10 of 17 published equivalence claims. The curation problem also remains open, as the strongest of the nine automated curators recovers only about half of the repairs. We emphasize that visual skill libraries should be curated by behavior, not by description. 
\end{abstract}
\vspace{-0.75cm}
\section{Introduction}
\label{sec:intro}

Behind every impressive vision-based agent system is an unglamorous dependence on a library of executable skills that must be maintained. These skills serve as executable vision tools that can power visual systems, such as VisProg~\cite{gupta2023visual}, Visual ChatGPT~\cite{wu2023visual}, MM-ReAct~\cite{yang2023mm}, Chameleon~\cite{lu2023chameleon}, and CLOVA~\cite{gao2024clova}. While the libraries are conventionally crafted and maintained by hand, \emph{can these agents autonomously maintain their vision-skill libraries as these libraries evolve over time?}

\begin{figure*}[ht]
    \centering
    \includegraphics[width=0.875\linewidth]{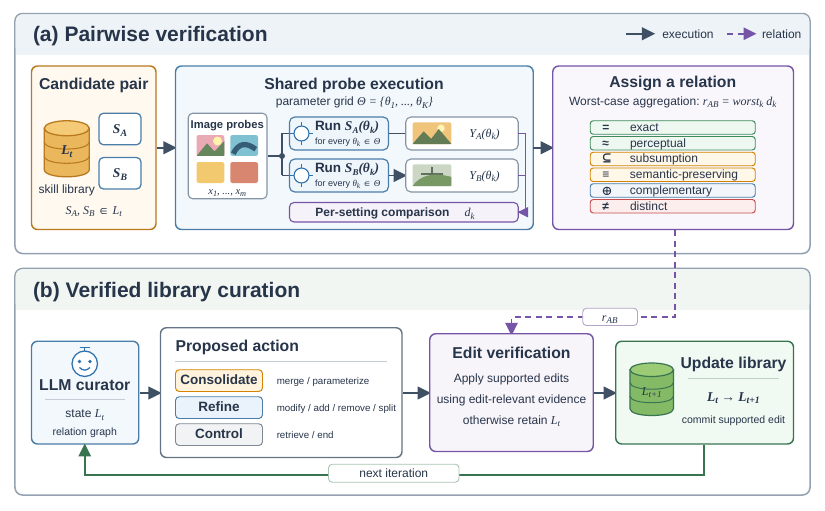}
    \caption{\textbf{\emph{Overview of the VisCurate pipeline:}} (a) A noisy seed skill library is evaluated through an output-grounded equivalence verifier, which executes candidate skills on shared probes and certifies their behavioral relations from the resulting outputs. (b) These verified relations guide the structural edits proposed by the curation agent, which are validated before updating the library. The process repeats until a compact and organized skill library is produced.}
    \label{fig:overview}
\end{figure*}

A visual skill is a deterministic function that transforms an input image into an output image, implemented with standard image-processing libraries such as OpenCV~\cite{bradski2000opencv}, PIL\footnote{\url{https://python-pillow.org}}, and NumPy~\cite{harris2020array}. Stochastic and neural operators are an extension of the same definition (\S\ref{sec:discussion}). An agent's effectiveness depends not only on its underlying model but also on the quality of its accessible skill library.\medskip

Maintaining a growing library is difficult as independent contributions introduce duplicate, near-duplicate, obsolete, and faulty skills. Repositories with large skill collections may contain redundancy, with 137,470 analyzed skills corresponding to approximately 56,043 distinct skill concepts~\cite{zhu2026skillclone}, yet prior work mostly focuses on skill acquisition~\cite{grand2024lilo,stengel2024regal,zheng2025skillweaver,alzubi2026evoskill,xu2026agent}. Curation hinges on deciding whether two skills are equivalent, and the signals normally used to decide that are unreliable. Similar descriptions can hide different outputs, causing a \emph{silent merge}, while different implementations can behave identically, causing a \emph{missed consolidation}~\cite{hu2026skillbrew,zhang2026skillcomposer,zhu2026skillclone}. We therefore compare the executed visual outputs instead of relying on descriptors.\medskip

We introduce VisCurate, a framework and benchmark for output-grounded visual-skill curation (\cref{fig:overview}, \cref{sec:method}). VisCurate executes candidate pairs on a shared probe battery and assigns relations that map onto safe library actions. We validate the framework itself at every layer by testing the central claim against an eight-rung ladder of name, embedding, description, and source-code baselines, and score the verifier against ground-truth relation graphs. On two real, uncurated libraries, the verifier finds most of the surface a package shares with its downstream copy to be exactly redundant, whereas published cross-library equivalence claims often fail at runtime. Output-derived screening turns quadratic pairing near-linear. However, curation itself remains hard as the strongest of nine language-model curators recovers only about half of the required repairs. Our contributions are:

\begin{itemize}
\medskip
    \item \textbf{Output-grounded visual equivalence.} A framework that relates two visual skills by executing both on shared probes, with a behavioral taxonomy in which each relation corresponds to one safe curation action.
    \item \textbf{Corruption benchmark.} We introduce seven types of defects into a clean reference library at configurable rates, creating 300 reproducible instances with known ground-truth relations and curation actions.\medskip
    \item \textbf{Framework validation.} A comparison against eight text, embedding, and source-code baselines, verifier self-validation against ground-truth graphs, audits of two real uncurated libraries, and a measured out-of-distribution certificate-flip rate.\medskip
    \item \textbf{Cost and curation studies.} A scalability analysis showing near-linear candidate growth with measured per-repair token costs, and an extensive evaluation of nine language-model curators showing that reliable and autonomous library repair remains an open problem.
\end{itemize}

\section{Related Work}
\label{sec:relatedwork}

\subsection{Skill Libraries for Visual Agents}

Compositional visual agents established that libraries of executable skills are the right substrate for complex tasks, \eg, VisProg~\cite{gupta2023visual} and ViperGPT~\cite{suris2023vipergpt} compose predefined vision and language modules into programs, CUA-Skill~\cite{chen2026cua} retrieves parameterized skills for graphical-interface tasks, VisualSkill~\cite{jiang2026visualskill} keeps visual content so computer-use agents can verify states, and PolySkill~\cite{yu2025learning} separates a skill's goal from its implementation to generalize better. All of these treat the library as fixed or append-only. A second line manages growth, \eg, Skill2Mem~\cite{anonymous2026skillmem} shows that uncurated public skills harm agent memory, the Agent Skills survey~\cite{xu2026agent} reports heavy redundancy in public collections, and clone detection~\cite{zhu2026skillclone}, SkillOS~\cite{ouyang2026skillos} and SkillComposer~\cite{zhang2026skillcomposer} find and manage skills through text, code, or downstream utility, while ContractSkill~\cite{lu2026contractskill}, GraSP~\cite{xia2026grasp}, Toolformer~\cite{schick2023toolformer}, and HuggingGPT~\cite{shen2023hugginggpt} repair, organize, or add individual tools. None of these methods can tell from text, code, or delayed rewards whether two visual skills apply the same transformation. \emph{VisCurate differs by deciding every library edit from executed outputs rather than from what a skill says about itself.}

\subsection{Program Equivalence and Library Learning}

Library-learning methods seek reusable abstractions and face the same equivalence question in symbolic form. DreamCoder~\cite{ellis2021dreamcoder} builds and compresses symbolic libraries, LILO~\cite{grand2024lilo} uses documentation to guide synthesis and compression, and ReGAL~\cite{stengel2024regal} refactors programs while checking that their outputs are preserved, making it the closest method to ours. All operate on domains where equivalence can be read from program structure or exact return values. \emph{VisCurate instead compares existing visual operations by their image outputs, where similar code or descriptions can still hide different behavior, and turns each verified relation into a concrete, safe library edit.}

\section{Methodology}
\label{sec:method}

The literature on agents and tool use has defined \emph{skill} in many ways, usually meaning the description of a \emph{workflow}~\cite{zhou2026comprehensive,liu2026skillrevise}. We instantiate a skill as a workflow over audited visual operations, each a single deterministic image-to-image function. Fixing the input and output modalities is what makes behavioral equivalence \emph{decidable by execution on a probe battery} rather than by text.

\subsection{Problem Formulation}

\textbf{Skills.} Let $\mathcal{I}=\bigcup_{H,W,C}[0,1]^{H\times W\times C}$ denote the space of images across resolutions and channel layouts. A skill is a tuple
$s=\langle k,\;\mu,\;d,\;f_s,\;\Theta_s\rangle$: an identifier $k$, a textual name $\mu$, a natural-language description $d$, a deterministic visual transformation $f_s:\mathcal{I}\times\Theta_s\to\mathcal{I}$, and a typed parameter schema $\Theta_s$ with concrete values $\theta\in\Theta_s$. A skill is retrieved via its text attributes $(\mu,d)$, and seeded-stochastic skills take the seed as part of the input, making them deterministic conditioned on it. A \emph{relation graph} $G=(L,E)$ labels each skill pair with a relation from the equivalence taxonomy (\S\ref{subsec:equivalenceTaxonomy}).

\medskip
\noindent\textbf{Environment and agent.} A corrupted library (\cref{sec:setup}) is the environment of a partially observable Markov decision process whose hidden state is the set of injected defects. The curation agent, in this work a frozen language model used as-is, is a policy over the observable history that proposes actions from a fixed set (\texttt{add}, \texttt{remove}, \texttt{modify}, \texttt{retrieve}, \texttt{merge}, \texttt{split}, \texttt{parameterize}, \texttt{end}). Every structural edit passes a verifier check that belongs to the environment rather than to the agent: an edit takes effect only if the output-grounded verifier certifies a relation that permits it. Episodes are scored against an ideal-action key derived from the injected defects, by an oracle reward the agent never observes, so the agent is the only thing that varies while the environment, the corruptions, and the checks stay fixed. 


\medskip
\noindent\textbf{Skill library and corruption.} A skill library is an ordered collection of skills
$L=(s_1,\dots,s_n)$, where $L\in\mathcal{L}=\bigcup_{n\ge 0}\mathcal{S}^{n}$. A log entry $\ell\in\mathcal{E}$ is the reproducible record a corruption emits when applied. It records the corruption defect $c$, the skills it altered $(s_k)_{k\in{\mathcal{K}}\subseteq\{1,\dots n\}}$, the configuration $\gamma\in\Gamma_c$ that produced it, the edits the defect induces on the relation graph, $\Delta G_\ell$, and the ideal curative action that repairs it, $a^*\in\mathcal{A}$. The ordered entries form the corruption log, from which every corrupted artifact and its ground truth can be replayed. We define
\begin{equation}
   \Delta(\log) = \Delta G_{\ell_n} \circ \dots \circ \Delta G_{\ell_1}
\end{equation}
as the aggregate edit the log induces on the graph, so that $G_\rho = G_0 \cup \Delta(\text{log})$. Each defect $c$ induces a function factory
$\mathsf{F}_c : \Gamma_c \to \bigl(\mathcal{L} \to \mathcal{L} \times \mathcal{E}\bigr)$. Given a config $\gamma$, made up of a target ID $k$, a seed $\omega$, and corruption
parameters $\theta_c$, it returns a corruption function $f_c = \mathsf{F}_c(\gamma)$ with
\begin{equation}
  f_c(L) = (L',\,\ell), \qquad L' \in \mathcal{L},\ \ \ell \in \mathcal{E},
\end{equation}
where $L'$ is the corrupted library. We note that, since $L'$ depends only on $L$ and $\gamma$, re-applying $f_c$ under the same seed $\omega$ reproduces $L'$ exactly.

\medskip
\noindent\textbf{Environment.} The library is the environment of the curation agent, formalized as a partially observable Markov decision process $\langle \mathcal{L},\mathcal{A},\mathcal{O},\Omega,{T},T_0,R\rangle$. The state $\mathbf{s}_t \in \mathcal{L}$ is the current (possibly corrupted) library $L_t$. The injected defects are part of the state but hidden from the agent, which is the source of partial observability. The action set $\mathcal{A}$ holds the library operations of \cref{tab:action_space}, $\mathcal{O}$ is the observation space, $\Omega : \mathcal{L} \to \mathcal{O}$ the observation function, and $T : \mathcal{L} \times \mathcal{A} \to \mathcal{L}$ the deterministic transition function. A non-structural action such as \texttt{retrieve} leaves $L_t$ unchanged and only surfaces information. A structural edit passes through a verifier check, which asks the output-grounded verifier for a certifying relation $r$ and applies the edit only if $r$ permits it, leaving the library unchanged otherwise. This check belongs to the environment, not to the agent. The action \textsc{end} moves to a terminal state. $T_0$ is the initial-state distribution, and $R : \mathcal{L} \times \mathcal{A} \to \{0,1\}$ is an oracle reward scored against the ideal action $a^*$ in the corruption log.

\medskip
\noindent\textbf{Curation agent.} The agent is a policy $\pi_\phi$ over the observable history
$h_t = (o_0, a_0, o_1, \dots, a_{t-1}, o_t) \in \mathcal{H}$,
\begin{equation}
  \pi_\phi : \mathcal{H} \to \Delta(\mathcal{A}), \qquad a_t \sim \pi_\phi(\,\cdot \mid h_t),
\end{equation}
choosing $a_t$ from the admissible actions. It maintains a belief
$\beta_t \in \Delta(\mathcal{L})$ over the hidden corruption and conditions its actions on it. Policies are compared by the return
\begin{equation}
    J(\pi_\phi) = \mathbb{E}_{\mathbf{s}_0 \sim T_0,\ a_t \sim \pi_\phi}
  \Bigl[\textstyle\sum_{t=0}^{H} R(\mathbf{s}_t, a_t)\Bigr],
\end{equation}
which is an evaluation objective. Here $\pi_\phi$ is a language model used as-is, so no policy is trained against $J$.

\subsection{The Equivalence Taxonomy}
\label{subsec:equivalenceTaxonomy}

We classify each ordered pair $(s_a,s_b)$ into exactly one of seven output-grounded relations. The classifier is \emph{hierarchical and stop-at-first} (Alg. \ref{alg:classify}). It evaluates relations cheapest-and-strictest first and returns the first that holds, so a stronger, more actionable relation is never masked by a weaker one. \textsc{COMPLEMENTARY} is the sole exception,  as it is an advisory annotation and does not affect edit decisions. The perceptual and semantic stages use pluggable learned backends: LPIPS \cite{lpips2018} for perceptual distance, and DINO ViT \cite{dinoViT} and CLIP \cite{clip-pmlr-v139-radford21a} feature spaces for semantics.

\begin{algorithm}[t]
\caption{Classifying an output pair $(s_a,s_b)$. $\mathsf{pc},\mathsf{sm}$ flag the
perceptual / semantic backends.}
\label{alg:classify}
\begin{algorithmic}[1]
\Require tolerances $\epsilon,\tau_{\mathrm{pc}},\tau_{\mathrm{sm}}$ and band $\delta$
\If{$D_{\mathrm{ex}}\le\epsilon$}
    \State \Return \textsc{Exact}
\ElsIf{$\mathsf{pc}\wedge D_{\mathrm{pc}}\!\le\!\tau_{\mathrm{pc}}\wedge\textsc{ssim}$}
    \State \Return \textsc{Perceptual}
\ElsIf{$\mathsf{pc}\wedge D_{\mathrm{pc}}\!\le\!\tau_{\mathrm{pc}}\!+\!\delta$}
    \State \Return \textsc{Uncertain}
\ElsIf{$s_a\sqsubseteq s_b \vee s_b\sqsubseteq s_a$}
    \State \Return \textsc{Subsumption}
\ElsIf{$\mathsf{sm}\wedge D_{\mathrm{sm}}\!\le\!\tau_{\mathrm{sm}}$}
    \State \Return \textsc{Semantic-Preserving}
\ElsIf{$\mathsf{sm}\wedge D_{\mathrm{sm}}\!\le\!\tau_{\mathrm{sm}}\!+\!\delta$}
    \State \Return \textsc{Uncertain}
\Else
    \State \Return \textsc{Distinct}
\EndIf
\end{algorithmic}
\end{algorithm}

\medskip
\noindent\textbf{Parameter grid.} Because most skills are parameterized, \textit{e.g.,} a blur skill has a kernel size, we never compare at a single operating point. We evaluate over a matched grid
$\Theta^{\mathrm{grid}}\subseteq\Theta_{s_a}\cap\Theta_{s_b}$ and aggregate by worst case
\begin{align}
D(s_a,s_b)=\max_{\theta\in\Theta^{\mathrm{grid}}}\;
\frac{1}{m}\sum_{i=1}^{m} d\!\Big(&\phi\big(f_{s_a}(x_i,\theta)\big),\,\nonumber\\
&\phi\big(f_{s_b}(x_i,\theta)\big)\Big),
\end{align}
where $\phi$ canonicalizes outputs and $x_i$ ranges over the probe battery (\S\ref{subsec:behaviorSignatureCanonicalization}). The worst case is what blocks spurious merges between skills that coincide on part of their domain: a Gaussian blur and a box blur agree at small kernels but diverge at large ones, and the max over the grid surfaces that divergence rather than averaging it away.\medskip

\noindent\textbf{Relations.}
(i)~\textbf{Exact}: outputs are identical across the battery at every shared parameter setting, up to a negligible $L_\infty$ tolerance.
(ii)~\textbf{Perceptual} relaxes exactness to human-indistinguishability, worst-case LPIPS below $\tau_{\mathrm{pc}}$ with a structural SSIM floor guarding LPIPS blind spots. Like \textsc{EXACT}, it permits a merge.
(iii)~\textbf{Subsumption} is directional: $s_a\sqsubseteq s_b$ holds when some setting of $s_b$ reproduces every setting of $s_a$ within perceptual tolerance, certified by searching $s_b$'s parameter schema (a fixed-angle rotation is subsumed by rotate-by-degrees).
(iv)~\textbf{Semantic-preserving} captures pairs that are neither pixel- nor perceptually-identical but occupy the same region of a learned feature space, \textit{e.g.,} two edge detectors, certified by DINO cosine distance (p90 cross-checked against the mean, robust to isolated feature-space outliers). It permits parameterization, never an outright merge.
(v)~\textbf{Complementarity} captures pairs that compose into a meaningful pipeline, \textit{e.g.,} dilation and erosion, requiring non-triviality of both skills and an executed composition-effect test; advisory only.
(vi)~\textbf{Distinct} is the default verdict by elimination, including spatial-extent mismatches with no subsumption certificate.
(vii)~\textbf{Uncertain} is an intentional abstention: a pair whose deciding distance clears a calibrated threshold too narrowly (within a band $\delta$) is routed to human review instead of being pushed into an unwarranted merge.

\begin{table}[ht]
\centering
\caption{Description of the action space of the curation agent.}
\label{tab:action_space}
\begin{tabular}{l p{0.60\columnwidth}}
\toprule
\textbf{Action} & \textbf{Meaning} \\
\midrule
\texttt{add}          & Add a new skill \\
\texttt{remove}       & Remove a skill \\
\texttt{modify}       & Modify metadata/schema or function behavior \\
\texttt{retrieve}     & Retrieve/query a skill \\
\texttt{merge}        & Merge redundant skills \\
\texttt{split}        & Split one skill into multiple skills \\
\texttt{parameterize} & Fold a specialization into a more general parameterized skill \\
\texttt{end}          & Stop the episode \\
\bottomrule
\end{tabular}
\end{table}

\subsection{Behavioral Signatures and Canonicalization}
\label{subsec:behaviorSignatureCanonicalization}
A probe battery $B=(x_1,\dots,x_m)$, with each $x_i\in\mathcal{I}$, spans input images, channel layouts (grayscale, RGB, RGBA), bit depths, and degenerate cases. For a skill $s$ evaluated at a parameter value $\theta\in\Theta_s$, the behavioral signature is
\begin{equation}
\sigma(s,\theta)=\big(\phi(f_s(x_1,\theta)),\,\dots,\,\phi(f_s(x_m,\theta))\big),
\end{equation}
where $\phi:\mathcal{I}\to\mathcal{C}$ maps outputs into a comparison space
$\mathcal{C}$. Equivalence is decided over signatures, so instead of ``are $s_a,s_b$ equivalent?'', we ask
``do $\sigma(s_a,\cdot)$ and $\sigma(s_b,\cdot)$ agree under a relation-specific metric?''.

Canonicalization $\phi$ removes representational noise, \textit{i.e.,} differences in channel layout, format, or encoding, so that distance reflects behavior rather than how an output happens to be stored. Two safeguards keep the comparison reliable. A shape check sets the distance $d=\infty$ whenever two outputs differ in spatial extent, so a resolution-changing skill is never merged with a resolution-preserving one. A content hash over the canonical form fires only for byte-identical outputs and never across shapes.

Three skill classes need care and are handled explicitly: (a) \emph{seeded-stochastic} skills
(\textit{e.g.,} additive noise) are deterministic at a fixed seed and are compared only at matched seeds, (b)
\emph{precision-sensitive} skills (\textit{e.g.,} FFT round-trips) are compared with perceptual rather than
exact tolerance, and (c) \emph{platform-sensitive} skills (\textit{e.g.,} seeded $k$-means quantization) are compared within-platform only. These classifications are recorded with the library and asserted in tests.

\section{Experimental Setup}
\label{sec:setup}

\subsection{Base Library} We hand-author a clean base library $L_0$ of 100 skills spanning geometric/canvas, photometric, signal/morphology, and mask/reconstruction families, with a designed ground-truth relation graph $G_0$ validated for DAG and transitive-closure consistency. $L_0$ embeds known relations across the classes curation depends on: directional subsumptions, complementary pairs, and engineered hard negatives (\textit{e.g.,} Gaussian vs.\ box blur, nearest vs.\ bilinear vs.\ bicubic resize) that must \emph{not} be merged. $G_0$ enumerates $944$ designed ordered pairs, is fixed before any metric runs and never derived from the metrics under test, and any unlisted pair is \textsc{DISTINCT} by default.

\subsection{Corruption} To emulate realistic library decay, we inject defects into $L_0$ at a controlled rate $\rho$ and composition $\boldsymbol{\alpha}$. Seven defect types are implemented as named, replayable function factories: (i) an \emph{implementation bug} that silently corrupts outputs, (ii) a \emph{metadata mislead} that rewrites name and description while leaving behavior intact, (iii) a \emph{duplicate} (exact or perceptual copy), (iv) a \emph{subsumption} (fixed-parameter specialization of a more general skill), (v) a \emph{parameter-schema drift}, (vi) a \emph{domain-scoped bug} that misbehaves only on some inputs, and (vii) a \emph{dead/identity skill}. A deterministic planner apportions the defect budget under a composition vector $\boldsymbol{\alpha}$ (\textsc{uniform}, \textsc{duplicate-heavy}, \textsc{metadata-heavy}), and a pure replay reconstructs the corrupted library $L_\rho$, its relation graph $G_\rho$, and an ideal-action key specifying the curative action each defect warrants. Corrupted skills masquerade as built-in, so a defect is observable only through behavior. Sweeping $\rho \in \{0.1,\dots,1.0\}$, the $3$ compositions, $5$ seeds, and $\text{mode}\in\{\text{single},\text{mixed}\}$ gives $300$ exactly reproducible instances.

\subsection{Curation Protocol} 
A curation agent works through one episode over a corrupted library $L_\rho$. It sees only redacted summaries of the skills and may propose up to $200$ actions. Code-bearing actions run inside the hardened sandbox so that their effects can be observed, and no structural edit reaches the library until the verifier has certified a relation supporting it. Actions the verifier turns down still cost the agent budget. Throughout the agent study, we run the verifier in its deterministic configuration with the learned backends switched off, so only \textsc{EXACT}, \textsc{SUBSUMPTION}, and \textsc{DISTINCT} can support an edit. This is the most conservative merge policy available to us, and it reaches the same decision every time, so all $2{,}700$ episodes replay exactly, and every agent sees the same $300$ instances.

\subsection{Equivalence Benchmark}
\label{sec:equivBenchmark}

Separately from the curation-agent study, we evaluate the output-grounded verifier \emph{directly} against a ladder of text and source-code baselines on the designed relation graph $G_0$, isolating the equivalence decision from the agent that consumes it. The metadata-mislead defect rewrites the text surface that every text baseline reads and that the text-blind verifier never sees. All numbers come from deterministic runs over the $100$-skill library, its $944$ designed ordered pairs, and the $177$-probe battery. Because only \textsc{EXACT} and \textsc{PERCEPTUAL} permit a merge, the binary \emph{mergeable} decision is the common ground on which every method can be compared directly.

\medskip
\noindent\textbf{Baselines.} We arrange eight baselines by how much of the skill they read: (1)~\emph{name-match} token overlap, (2)~\emph{TF--IDF cosine}, (3)~\emph{sentence-embedding cosine} over name and description, (4)~\emph{code-embedding cosine} over source, (5a/5b)~\emph{AST similarity} in structure-only and semantics-preserving variants, (6)~an \emph{LLM judge reading descriptions}, (7)~an \emph{LLM judge reading source}, and (8)~the \emph{output-grounded verifier} (\texttt{lpips-alex}, DINO ViT-B/16, CLIP ViT-B/32 backends). Each rung emits a mergeable decision at a matched operating point, \eg, the lowest threshold keeping precision on non-equivalence at $\ge 0.99$ with merge recall $\ge 0.5$, falling back to defaults where that rule cannot bind (full rung definitions in \S\ref{sec:app_ladder}). Because a clean reference graph designates no true duplicates, $G_0$ supports the safety comparison only, and merge \emph{recall} is measured against the corrupted graphs $G_\rho$ (\S\ref{subsec:selfval}).

\medskip
\noindent\textbf{Metrics.} In curation, a wrong merge is far more costly than a missed one. We lead this principle with \emph{safety on non-equivalence}, which is the false-merge rate on truly-\textsc{DISTINCT} pairs, reported over all distinct pairs and separately over the hard-negative slice. Per-relation precision, recall, and F1 then quantify how well a method recovers the taxonomy. Curation-agent metrics are defined in \S\ref{sec:app_metrics}.

\begin{table}[ht]
\centering
\caption{The baseline ladder on the $944$ designed pairs of $G_0$ at matched operating points. FPR$_{\mathrm{dist}}$: false-merge rate on truly-\textsc{DISTINCT} pairs, in percent. FM$_{\mathrm{hard}}$: false merges accepted on the six engineered hard negatives. VisCurate abstains on $1.5\%$ of decisions; no baseline abstains. The source-reading LLM judge is scored on a 120-pair subsample (mean of three repeats), so its FPR is not directly comparable and is omitted. Exact 95\% CIs for every count are in the appendix \S\ref{sec:app_ladder}.}
\label{tab:verifier-safety}
\renewcommand{\arraystretch}{1.08}
\resizebox{0.975\columnwidth}{!}{%
\begin{tabular}{@{}llcc@{}}
\toprule
\textbf{Method} & \textbf{Reads} & \textbf{FPR$_{\mathrm{dist}}$ (\%)}\,$\downarrow$ & \textbf{FM$_{\mathrm{hard}}$}\,$\downarrow$ \\
\midrule
Name match          & name        & \cellcolor{red!20}5.40   & \cellcolor{red!25}5 \\
TF--IDF cosine      & description & \cellcolor{green!12}0.21 & \cellcolor{green!12}0 \\
Sentence embedding  & description & \cellcolor{red!20}5.61   & \cellcolor{red!25}5 \\
Code embedding      & source      & \cellcolor{orange!15}1.80 & \cellcolor{red!18}4 \\
AST (structural)    & source      & \cellcolor{green!12}0.00 & \cellcolor{green!12}0 \\
AST (semantic)      & source      & \cellcolor{green!12}0.00 & \cellcolor{green!12}0 \\
LLM judge           & description & \cellcolor{green!12}0.11 & \cellcolor{orange!15}1 \\
LLM judge           & source      & --                        & \cellcolor{green!12}0 \\
\midrule
\rowcolor{cyan!07}
\textbf{VisCurate (ours)} & \textbf{outputs} & \textbf{0.00} & \textbf{0} \\
\bottomrule
\end{tabular}%
}
\end{table}
\section{Experiments and Results}
\label{sec:results}

We organize the evaluation around five questions, one per layer of the framework.
\textbf{RQ1:} Do names, descriptions, or source code tell us how a skill behaves (\S\ref{subsec:verifResults})?
\textbf{RQ2:} How many of the truly redundant pairs does the verifier find, and how often does it get one wrong (\S\ref{subsec:selfval})?
\textbf{RQ3:} Does that result hold on a larger set of hard negatives (\S\ref{subsec:hardneg})?
\textbf{RQ4:} Does the framework carry over to real, uncurated libraries (\S\ref{subsec:real})?
\textbf{RQ5:} How well do current automated pipelines using LLM-based agents curate a decaying library (\S\ref{subsec:agents}), and what does it cost (\S\ref{subsec:cost})?
\subsection{Text and Code do not predict Visual Behavior}
\label{subsec:verifResults}

If names, descriptions, or source code told us how a skill behaves, a matcher built on them should make safe merge decisions on $G_0$'s $944$ designed pairs. Since $G_0$ designates no true duplicates, every merge proposed on it is false by construction. \cref{tab:verifier-safety} reports each method's false-merge rate on truly-\textsc{DISTINCT} pairs and false-merge count on the six engineered hard negatives.

\medskip
\noindent\textbf{Surface similarity over-merges.} Name matching and sentence embeddings each accept five of the six possible false merges on the hard negatives, and the code embedding accepts four. The output-grounded verifier accepts none anywhere on the graph, and abstains on $1.5\%$ of decisions.

\medskip
\noindent\textbf{Stronger baselines stay structurally blind.} The description-reading LLM judge merged one hard negative in this run, and a rerun with a newer judge produced zero, so we rest no claim on strong judges over-merging. However, every method except ours emits only a merge-vs-distinct split, so none can represent the subsumption direction, semantic preservation, or complementarity, and no text method is, in principle, sensitive to the metadata-mislead defect. On a six-pair slice, a perfect $0/6$ still bounds the false-merge rate only at $39\%$ (one-sided 95\%), which is why RQ3 expands the slice by an order of magnitude.

\medskip
\noindent\textbf{Source code fails in both directions at once.} Duplicates enter libraries as thin delegating wrappers. A wrapper executes identically to its donor yet shares almost no source text with it, while two wrappers of \emph{unrelated} skills are nearly identical as source. On one fully corrupted instance, $25.7\%$ of cross-original clone pairs are byte-identical in source while wrapping different skills, so a source-similarity method is wrong both ways at once; output grounding is immune by construction (\S\ref{sec:app_ladder}).

\medskip
\noindent\textbf{Fine-grained taxonomy.} Only the verifier resolves the full six behavioral relation taxonomy (\cref{fig:verifier-relations}). \textsc{SUBSUMPTION}, the relation behind parameterization, is recovered well (F1 $0.750$). \textsc{SEMANTIC-PRESERVING} is the weak point (precision $0.037$), though it is not a safety failure. The distinct pairs it misses are labels that permit no merging.

\subsection{Redundancy recovered by the Verifier}
\label{subsec:selfval}

The corrupted graphs $G_\rho$ carry known relations by construction, including injected duplicates. So they support what $G_0$ cannot, \ie, measuring how much real redundancy the verifier recovers. We replay the verifier over all 300 corruption instances, scoring $13{,}658$ labeled pairs with $3{,}310$ truly mergeable positives (\textsc{EXACT} $\cup$ \textsc{PERCEPTUAL}), zero unresolved execution errors, and zero verdict disagreements across re-run shards (\S\ref{sec:app_selfval}).

\begin{table}[t]
\centering
\caption{Verifier self-validation against ground-truth graphs $G_\rho$ (300 instances, $13{,}658$ pairs). The binary merge decision scores recall $0.512$ (exact 95\% CI $[0.495, 0.530]$) at precision $1.000$ (tp/fp/fn $= 1{,}696/0/1{,}614$).}
\label{tab:selfval}
\resizebox{0.97\columnwidth}{!}{%
\begin{tabular}{lccc}
\toprule
\textbf{Relation} & \textbf{Support} & \textbf{Recall} & \textbf{Precision} \\
\midrule
\textsc{Exact}               & 1{,}640 & 0.968 & 1.000 \\
\textsc{Perceptual}          & 1{,}670 & 0.046 & 0.704 \\
\textsc{Subsumption}         & 4{,}672 & 0.758 & 0.996 \\
\textsc{Semantic-Preserving} & 3{,}424 & 0.319 & 0.278 \\
\textsc{Complementary}       & 310 & 0.000 & 0.000 \\
\textsc{Distinct}            & 1{,}942 & 0.384 & 0.188 \\
\bottomrule
\end{tabular}%
}
\end{table}

\medskip
\noindent\textbf{Safety holds everywhere.} The verifier finds $51.2\%$ of the pairs that are truly mergeable and makes \emph{no} false merges (Tab. \ref{tab:selfval}). That count stays at zero at each corruption rate $\rho \in \{0.1,\dots,1.0\}$ taken on its own (per-rate table in \S\ref{sec:app_selfval}), so safety does not decay as the library gets dirtier, and a 34-instance pilot agreed (recall $0.493$).

\medskip
\noindent\textbf{The misses sit one threshold away.} Byte-identical duplicates are found almost always (\textsc{EXACT} recall $0.968$) and near-duplicates almost never (\textsc{PERCEPTUAL} recall $0.046$), because $1{,}594$ of $1{,}670$ \textsc{PERCEPTUAL} pairs are labeled \textsc{SEMANTIC-PRESERVING}, one conservative notch below the label that would allow their merge.

\subsection{Scaling Hard-Negatives}
\label{subsec:hardneg}

The six designed hard negatives support only a weak bound, so we expand the slice to $69$ engineered pairs spanning the same trap families (resampling kernels, border modes, quantization variants). The verifier makes 0/69 false merges, and pairs checked independently by more than one shard agree in every case (\textit{c.f.} \S\ref{sec:app_hardneg}).

\begin{table*}[ht]
\centering
\caption{Performance of nine curation agents on the corrupted-library benchmark. Every cell aggregates the same 300 instances, with instance-clustered bootstrap 95\% CIs (2{,}000 resamples). $\mu$P, $\mu$R, $\mu$F1: mean per-episode precision, recall, and F1. mF1: micro F1. IS: Intrinsic Score. AC: action cost. CR: compression ratio. Comp.: mean compression. Pred.: mean predicted actions. Higher CR, Comp., and Pred.\ are not better on their own and carry no arrow, since they are only meaningful alongside precision or F1. Metric definitions are in \S\ref{sec:app_metrics}. Per-$\rho$ breakdowns and action outcomes are in \S\ref{sec:app_agents}.}
\label{tab:model_evaluation}
\begin{tabular}{lccccccccc}
\toprule
Model         & $\mu$P$\uparrow$ & $\mu$R$\uparrow$ & $\mu$F1$\uparrow$ [95\% CI] &   mF1$\uparrow$ &    IS$\uparrow$ &     AC$\downarrow$ &      CR &   Comp. &   Pred. \\
\midrule
Llama 3.2 1B \cite{meta2024llama32}  & 0.000 & 0.000 & 0.000 [0.000, 0.000] & 0.000 & 0.000 & 152.11 & $-0.001$ & $-0.11$ &  0.01 \\
Llama 3.2 3B \cite{meta2024llama32}  & 0.443 & 0.013 & 0.024 [0.019, 0.028] & 0.018 & 0.012 & 200.00 & 0.008 &  1.05 &  1.05 \\
Llama 3.1 8B \cite{grattafiori2024llama3}  & 0.615 & 0.061 & 0.095 [0.085, 0.106] & 0.070 & 0.049 & 200.00 & 0.030 &  3.67 &  4.08 \\\midrule
Qwen 3.5 2B \cite{qwen2026qwen35}   & 0.134 & 0.008 & 0.013 [0.009, 0.018] & 0.006 & 0.007 & 128.35 & 0.005 &  0.57 &  0.59 \\
Qwen 3.5 4B \cite{qwen2026qwen35}   & 0.290 & 0.098 & 0.109 [0.097, 0.122] & 0.089 & 0.063 & 185.37 & 0.098 & 11.48 & 11.48 \\
Qwen 3.5 9B \cite{qwen2026qwen35}  & 0.357 & 0.074 & 0.101 [0.085, 0.117] & 0.065 & 0.060 & 171.52 & 0.034 &  3.92 &  3.92 \\
Qwen 3.5 27B \cite{qwen2026qwen35}  & 0.662 & 0.423 & 0.505 [0.490, 0.521] & 0.515 & 0.358 & 109.59 & 0.222 & 29.31 & 37.90 \\\midrule
Gemma 4 12B  \cite{gemmateam2026gemma4}  & \cellcolor{cyan!10}\textbf{0.777} & 0.204 & 0.290 [0.275, 0.306] & 0.266 & 0.159 & 172.23 & 0.103 & 13.02 & 13.15 \\\midrule
GPT-5.5 \cite{openai2026gpt55}      & 0.735 & \cellcolor{cyan!10}\textbf{0.531} & \cellcolor{cyan!10}\textbf{0.587} [0.572, 0.601] & \cellcolor{cyan!10}\textbf{0.605} & \cellcolor{cyan!10}\textbf{0.496} &  \cellcolor{cyan!10}\textbf{78.86} & 0.130 & 16.47 & 41.42 \\
\bottomrule
\end{tabular}%
\end{table*}

\subsection{Performance on Real Libraries}
\label{subsec:real}

Synthetic corruption provides ground truth but not external validity, so we take the verifier, thresholds frozen (\cref{tab:thresholds}), to two real, uncurated libraries.

\medskip
\noindent\textbf{Cross-version.} We take two versions of the same codebase: \texttt{pilgram 2.0.0}\footnote{\url{https://github.com/akiomik/pilgram/tree/v2.0.0}} and \texttt{pilgram 2.0.10}\footnote{\url{https://github.com/mgineer85/pilgram2/tree/v2.0.10}}. They share 25 filter names, and neither package has methods to find overlaps. Of the 25 shared filters, 23 are byte-identical on all 177 probes and certified \textsc{EXACT}, so $92\%$ of the shared surface can be merged. The other two have really diverged, and the verifier correctly declines to merge them. The two copies differ visibly (LPIPS $0.246$) yet are semantically indistinguishable (DINO p90 distance $0.0008$), so a perceptual-only test would call it divergence while a semantic-only test merges it.

\medskip
\noindent\textbf{Cross-library.} Corpus A holds 93 implementations of common operations across 20 families from widely used packages (OpenCV, PIL, scikit-image, torchvision, Kornia, Albumentations), giving 182 within-family pairs, with all 93 passing the self-pair check. Published documentation tables group 17 of these pairs as directly equivalent, yet execution certifies only 10 of 17 ($0.588$, exact 95\% CI $[0.33, 0.82]$) as mergeable at aligned parameters, or 10 of 14 among fully alignable pairs. We read this as documentation grouping operations by category rather than by behavior, and not as a maintainer error. It is also the situation that motivates output grounding. While independent reimplementations of a single operation drift apart, copied packages remain identical. In neither case does anyone run the two side by side.

\medskip
\noindent\textbf{Oracle for real libraries.} Because Corpus A has no designed answer key, we build a mechanical differential oracle (dense parameter sweeps with representation-derived tolerances, \S\ref{sec:app_real}) that labels all 182 pairs with zero ambiguous cases. The oracle is the independent ground truth for the text and code baselines. It never executes anything, only performs an exhaustive-versus-approximate check for our verifier, and it is labeled as such wherever it appears. Its labels ship with the benchmark, which gives later matching methods something to score precision and recall against on real libraries rather than on synthetic graphs alone (\S\ref{sec:app_ladder}).

\subsection{Performance of LLM Agent Curation}
\label{subsec:agents}

We evaluate nine language-model curators from four families: Llama 3.1 and 3.2, Qwen 3.5, and Gemma 4 at multiple sizes as open-source models, and GPT-5.5 as the closed-source representative. Inference settings are reported in \S\ref{sec:app_training}, and metrics in \S\ref{sec:app_metrics}.

\begin{figure}[t]
\centering
\begin{tikzpicture}
\definecolor{vcBlue}{HTML}{0072B2}
\definecolor{vcGreen}{HTML}{009E73}
\definecolor{vcOrange}{HTML}{D55E00}
\definecolor{vcPink}{HTML}{CC79A7}
\begin{axis}[
    width=\columnwidth, height=4.8cm,
    xlabel={Corruption rate $\rho$},
    ylabel={Mean F1},
    xmin=0.05, xmax=1.05, ymin=0, ymax=0.72,
    xtick={0.1,0.2,0.3,0.4,0.5,0.6,0.7,0.8,0.9,1.0},
    xticklabel style={/pgf/number format/fixed},
    ymajorgrids=true,
    grid style={dashed, gray!25},
    tick label style={font=\footnotesize},
    label style={font=\small},
    legend style={font=\footnotesize, draw=none, fill=none,
                  at={(0.5,1.02)}, anchor=south, legend columns=2},
    legend cell align=left,
]
\addplot[vcBlue, thick, mark=*, mark size=1.6pt] coordinates
  {(0.1,0.420) (0.2,0.521) (0.3,0.579) (0.4,0.596) (0.5,0.621) (0.6,0.641) (0.7,0.624) (0.8,0.628) (0.9,0.625) (1.0,0.612)};
\addlegendentry{GPT-5.5}
\addplot[vcGreen, thick, mark=square*, mark size=1.5pt] coordinates
  {(0.1,0.457) (0.2,0.476) (0.3,0.522) (0.4,0.559) (0.5,0.523) (0.6,0.512) (0.7,0.524) (0.8,0.519) (0.9,0.484) (1.0,0.474)};
\addlegendentry{Qwen 3.5 27B}
\addplot[vcOrange, thick, mark=triangle*, mark size=1.9pt] coordinates
  {(0.1,0.378) (0.2,0.384) (0.3,0.356) (0.4,0.353) (0.5,0.321) (0.6,0.252) (0.7,0.231) (0.8,0.209) (0.9,0.210) (1.0,0.206)};
\addlegendentry{Gemma 4 12B}
\addplot[vcPink, thick, mark=diamond*, mark size=1.9pt] coordinates
  {(0.1,0.216) (0.2,0.226) (0.3,0.185) (0.4,0.160) (0.5,0.120) (0.6,0.068) (0.7,0.048) (0.8,0.033) (0.9,0.017) (1.0,0.018)};
\addlegendentry{Qwen 3.5 4B}
\end{axis}
\end{tikzpicture}
\caption{Mean F1 by corruption rate $\rho$ for the four strongest curators ($n{=}30$ episodes per cell). Difficulty is not monotone in $\rho$ as GPT-5.5 improves as defects multiply, Qwen 3.5 27B stays flat, and the smaller models collapse. Full results in \cref{tab:app_perrho}.}
\label{fig:f1-by-rho}
\end{figure}
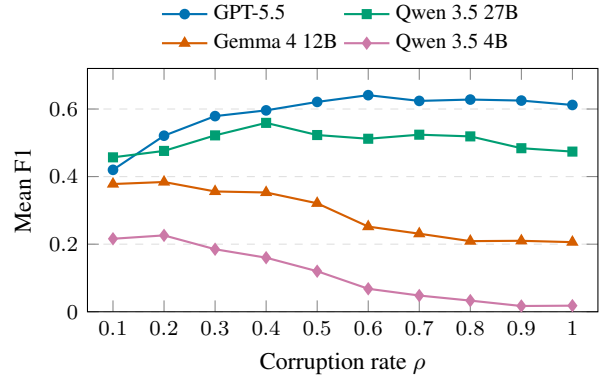

\medskip
\noindent\textbf{The best model recovers about half the repairs.} GPT-5.5 leads on recall ($0.531$), mean F1 ($0.587$), and Intrinsic Score ($0.496$), and it does so while attempting the fewest actions per episode, most of which are productive (\cref{tab:model_evaluation}). Qwen 3.5 27B is the only open model near it, with the best precision-recall balance in its tier and the largest real compression (\cref{tab:model_evaluation}). Gemma 4 12B is the most precise model in the study ($0.777$) but acts too rarely, repairing only a fifth of the faults. Performance scales with size within every family, and the smallest models do very little, \eg, Llama 3.2 1B never produces a correct action.

\medskip
\noindent\textbf{Difficulty does not simply rise with the corruption rate.} \cref{fig:f1-by-rho} shows mean F1 across the corruption sweep, and which way a model moves depends on the model. GPT-5.5 \emph{rises} from $0.420$ at $\rho{=}0.1$ to $0.641$ at $\rho{=}0.6$ and holds there, Qwen 3.5 27B stays flat, and Gemma 4 12B and Qwen 3.5 4B fall by roughly $0.18$. A capable curator gets more scorable repair opportunities as defects multiply, while a weaker one is simply overwhelmed by them.

\medskip
\noindent\textbf{Spending the budget is not the same as using it.} Llama 3.2 3B and 3.1 8B burn through all 200 actions and still predict only $1-4$ repairs, while the strongest models reach far higher prediction counts at a lower action cost. Intrinsic Score is below the mean F1 for every model, because much of the budget goes to actions the library never accepts.

\subsection{Cost and Scalability}
\label{subsec:cost}

Running skills and comparing their outputs are computationally expensive. We measure where that cost goes and how it grows, using $262{,}833$ telemetry events.

\medskip
\noindent\textbf{Screening makes pairing near-linear.} A cheap output-derived fingerprint proposes candidate pairs before the verifier runs. Sweeping the screen's $(k,\mathrm{radius})$ grid over ten corrupted instances shows that the neighbor count $k$ is what matters, and the radius almost never does. Recall of mergeable pairs reaches $1.000$ at $k{=}10$ while still discarding $95.1\%$ of all pairs (\S\ref{sec:app_cost}). The sweep also told us that our original default of $k{=}5$ was a poor choice, since it dropped $3.75\%$ of mergeable pairs and bought nothing for it, so we now use $k{=}10$. Across nested library subsamples, the candidate count fits $|C| = 2.38\,n^{1.13}$ ($R^2 = 0.998$) against the exponent $2$ of exhaustive pairing; extrapolated to $n{=}10^4$ (beyond the fit's support), screening proposes about $7.6{\times}10^4$ candidates instead of $5.0{\times}10^7$ exhaustive pairs.

\medskip
\noindent\textbf{Where the time and memory go.} The stop-at-first cascade settles $54.9\%$ of pairs before any learned backend runs, and the signature cache serves $78.2\%$ of executions. A pair takes $401$\,ms to decide at the median, and the p95 of $64.5$\,s comes almost entirely from subsumption searches. Memory constrains a deployment more than time does, as caching raw output sets costs $214.8$\,MB per skill and reaches $2.1$\,TB at $10^4$ skills, while caching only backend features costs $5.4$\,GB at the same scale, a $395\times$ saving. Evicting from that cache changes the wall-clock but never a verdict (\S\ref{sec:app_cost}).

\medskip
\noindent\textbf{Cost per correct repair.} On identical instrumented episodes, GPT-5.5 spends $9{,}202$ tokens per correct repair, with numerator and denominator from the same 41 telemetry-instrumented episodes, since mixing scopes distorts the ratio by up to $5\times$ (\S\ref{sec:app_cost}).

\section{Discussion and Limitations}
\label{sec:discussion}

VisCurate separates two questions: whether two skills behave alike, and whether the library should change. The verifier answers the first from the executed outputs, the agent answers the second, and an edit goes through only once the verifier certifies the required relation. We measure the verifier's safety rather than assert it, and it made zero false merges everywhere we looked. The verifier's caution costs about half of the mergeable pairs, and the curation agents remain far from solving the task.

Two skills with identical outputs can still differ in compute cost, dependency weight, and maintainability, and a malicious skill can match a benign one on every probe while carrying a harmful payload. Cost and maintainability are complementary inputs that the agent can weigh alongside the relation certificate, and side effects belong to the sandbox, which confines execution and blocks unauditable code. Output comparison also sees a domain-scoped bug only when the probes cover that domain, and it cannot flag a misleading description or an unused but correct skill. 

Several extension studies of our work have also been covered in \S\ref{sec:app_extensions} on neural and stochastic skills, out-of-distribution probes, and probe battery ablation.

\section{Conclusion}

We introduce VisCurate, an output-grounded framework for curating executable visual skill libraries. A metadata-blind verifier certifies behavioral relations, and a taxonomy translates them into safe edits. The verifier makes zero false merges, while surface baselines produce up to five of six, and the strongest of nine curators repairs only about half the faults. Behavioral equivalence requires execution, and VisCurate provides both the verifier and the benchmark. Our work reemphasizes that visual skills should be curated by their visual behavior, not by text descriptions. VisCurate also opened several new directions to improve the performance of curation agents on visual skill libraries.
\section*{Acknowledgements}
We thank Didhiti Nahid for her helpful feedback on an earlier draft of this paper. Md Farhan Ishraq, in particular wants to thank his parents, Md. Shah Alam and Afia Khatun Mily, for everything. 
{
  \small
  \bibliographystyle{ieeenat_fullname}
  \bibliography{main}
}

\clearpage
\setcounter{page}{1}
\maketitlesupplementary
\appendix
\renewcommand{\thefigure}{A.\arabic{figure}}
\setcounter{figure}{0}
\renewcommand{\thetable}{A.\arabic{table}}
\setcounter{table}{0}

\section{Additional Methodology Details}
\label{sec:app_method}

\subsection{Threshold Calibration}
\label{subsec:thresholdCalibration}

The thresholds $\epsilon$, $\tau_{\mathrm{pc}}$, $\tau_{\mathrm{sm}}$, and the band width $\delta$ are learned from a labeled calibration set containing $713$ pairs from eight operator families. We select the thresholds to maximize precision on \textsc{DISTINCT} decisions while maintaining a minimum recall for true equivalences. This favors keeping skills separate when the evidence for merging them is insufficient.

The calibration and evaluation sets are disjoint at the skill-family level, preventing the same family from appearing in both. We further check this separation using $100$ held-out pairs from additional families. The calibration split, along with its hash and date, is recorded before evaluation begins. Once calibrated, the thresholds remain fixed across all experiments, with no tuning on evaluation data (\cref{tab:thresholds}). Across separate executions of the full benchmark, the verifier produces matching relation labels for $943$ of the $944$ pairs.

\begin{table}[t]
\centering
\caption{Operating thresholds for the output-grounded verifier, calibrated before evaluation and held fixed across all experiments.}
\label{tab:thresholds}
\resizebox{\columnwidth}{!}{%
\begin{tabular}{l c l}
\toprule
\textbf{Threshold} & \textbf{Value} & \textbf{Role} \\
\midrule
$\epsilon$ (exact $L_\infty$)      & $1/255$ & numerical-equality tolerance \\
$\tau_{\mathrm{pc}}$ (LPIPS)       & $0.05$ & worst-case perceptual bound \\
SSIM floor ($1-$SSIM)              & $0.10$ & LPIPS blind-spot guard \\
$\tau_{\mathrm{sm}}$ (DINO, p90)   & $0.15$ & semantic-preserving bound \\
complementary (LPIPS)              & $0.05$ & composition-effect tolerance \\
abstention $\delta$                & $0.05$ & \textsc{UNCERTAIN}-band width \\
\bottomrule
\end{tabular}%
}
\end{table}

\subsection{Candidate Generation}
\label{sec:app_candidates}

As running the full comparator stack over all $\binom{n}{2}$ pairs is wasteful, we compute a cheap, purely output-derived \emph{fingerprint} per skill, a low-dimensional summary of its signature that combines a perceptual hash with a mean DINO feature over a small screening sub-battery. Pairs with fingerprints within a relatively generous distance are passed to the full hierarchical classifier. We also retain all same-family pairs and engineered hard negatives, regardless of their fingerprint distance, to ensure that these cases are evaluated directly. Fingerprints are computed from outputs, never from the skill's text $(\mu,d)$, so two skills with unrelated descriptions but identical behavior still collide and get verified, and the modality boundary holds end-to-end, including at candidate generation.

\subsection{Curation Under Output Checks}
\label{sec:outputCuration}

Curation operates on a redacted view of the skill library, and structural edits never take effect directly. Each edit passes through a \textbf{check} $\Gamma$ that asks the verifier for a certifying relation $r$ and then applies a fixed relation-to-action policy. \textsc{EXACT} and \textsc{PERCEPTUAL} permit \texttt{merge} into a single canonical skill. \textsc{SUBSUMPTION} in the correct direction  permits \texttt{parameterize}. \textsc{SEMANTIC-PRESERVING} permits \texttt{parameterize} or \texttt{merge}. \textsc{COMPLEMENTARY} and \textsc{DISTINCT} permit no structural edit, and \textsc{UNCERTAIN} goes to human review.

An edit that no certifying relation supports is \emph{rejected} with a structured reason carrying the relation, the deciding distances, and admissible alternatives. Rejected or blocked edits still consume budget. Usage acts as a secondary constraint, so a heavily used skill resists destructive consolidation even when a relation would allow it. A separate \textbf{trust boundary} governs agent-authored code. Any action that introduces or edits a skill's function, such as \textsc{split} or an \textsc{add}/\textsc{modify} that supplies source, runs only inside a hardened sandbox that isolates the child process. Only code that clears the sandbox can be output-verified, and anything the sandbox cannot admit is blocked and consumes budget without changing the library. 

\section{Experimental Setup Details}
\label{sec:app_setup}

\subsection{Inference and Serving}
\label{sec:app_training}

No model is trained or fine-tuned in this work. We evaluate each curator in its original, frozen form, allowing us to compare their curation behavior under the same experimental conditions. The evaluation includes open-weight models (Llama 3.1/3.2, Qwen 3.5, and Gemma 4) and hosted models (GPT-5.5 and GPT-5.6-luna). Open-weight models are served locally using vLLM 0.24, while hosted models are accessed through their provider APIs. The environment, corruption instances, and verification procedure remain unchanged across experiments, so differences in curation behavior can be attributed to the curator rather than changes in the evaluation setup.

Each episode has a budget of 200 actions and runs over the same 300 corruption instances. When an experiment is resumed, completed instances are skipped rather than evaluated again. The verifier's learned backends run separately from model inference, and the calibrated thresholds from \cref{tab:thresholds} are loaded in read-only mode. The LLM judges used in the comparison ladder (\cref{sec:app_ladder}) are queried at temperature 0 with a 512-token reply limit to keep their responses concise and consistent.

\subsection{Curation-Agent Metrics}
\label{sec:app_metrics}

\noindent\textbf{Mean Precision ($\mu$P):} average per-episode fraction of predicted repair actions that were correct.

\noindent\textbf{Mean Recall ($\mu$R):} average per-episode fraction of required repair actions the model found.

\noindent\textbf{Mean F1 ($\mu$F1):} average per-episode harmonic mean of precision and recall.

\noindent\textbf{Micro F1 (mF1):} F1 after pooling all true positives, false positives, and false negatives across episodes. This weights larger and harder episodes more than mean F1.

\noindent\textbf{Action Cost (AC):} number of non-\textsc{end} actions attempted per episode.

\noindent\textbf{Compression Ratio (CR):} fractional library shrinkage per episode. If the episode ends at step $T$, $\mathrm{CR}=(|L_0|-|L_T|)/|L_0|$. Higher means the curated library got smaller, which is only good alongside high precision or F1.

\noindent\textbf{Mean Compression (Comp.):} raw library-size reduction $|L_0|-|L_T|$.

\noindent\textbf{Mean Predicted (Pred.):} number of predicted repair actions per episode. Very low values indicate under-acting and very high values can indicate over-acting.

\noindent\textbf{Intrinsic Score (IS):} mean F1 discounted for budget wasted on non-productive actions. With $\mathrm{AC}$ the number of non-\textsc{end} actions and $u$ the number of those that were rejected, blocked, or invalid,
\begin{equation}
\mathrm{IS} \;=\; \mathrm{F1}\cdot\max\!\Big(0,\; 1-\tfrac{u}{2\,\mathrm{AC}}\Big),
\end{equation}
clipped to $[0,1]$ and set to $0$ when $\mathrm{AC}=0$. The penalty keeps the score independent of downstream query success while punishing policies that spend budget without improving the library.

\section{Verifier Validation Details}
\label{sec:app_validation}

\subsection{Baseline Ladder}
\label{sec:app_ladder}

\noindent\textbf{Rung definitions.} \emph{Name-match} takes the Jaccard token overlap over the union of name and identifier tokens and merges when the overlap is at least $0.5$, acting as a renamed-duplicate detector. \emph{TF--IDF cosine} embeds the concatenated name and description with deterministic TF--IDF (smoothed IDF, $L_2$-normalized) and merges at cosine $\ge 0.6$. \emph{Sentence-embedding cosine} replaces TF--IDF with a pretrained sentence encoder at the same matched threshold. \emph{Code-embedding cosine} embeds the skill source with a pretrained code encoder and merges at cosine $\ge 0.9$. The \emph{AST rungs} compare abstract syntax trees at $\tau=0.9$ in two variants. The structure-only variant canonicalizes literals and is therefore blind to parameter-default drift by construction, while the semantics-preserving variant normalizes only meaning-preserving rewrites and is the fairest source-code baseline. The \emph{LLM judges} are fixed, strong instruction-tuned models queried with reasoning disabled for one-word determinism. Each sees either both descriptions or both source bodies and must return exactly one relation word, and unparseable replies are treated conservatively as \textsc{DISTINCT}. The description judge scores all $944$ pairs, while the source judge scores a $120$-pair stratified subsample with three repeats for cost reasons. Both judges are deliberately distinct from every curation-subject model. The matched operating point for each threshold rung is the lowest threshold keeping precision on non-equivalence at $\ge 0.99$ with merge recall $\ge 0.5$, falling back to the shipped default where the rule cannot bind.


\begin{figure*}
    \centering
    \includegraphics[width=0.75\linewidth]{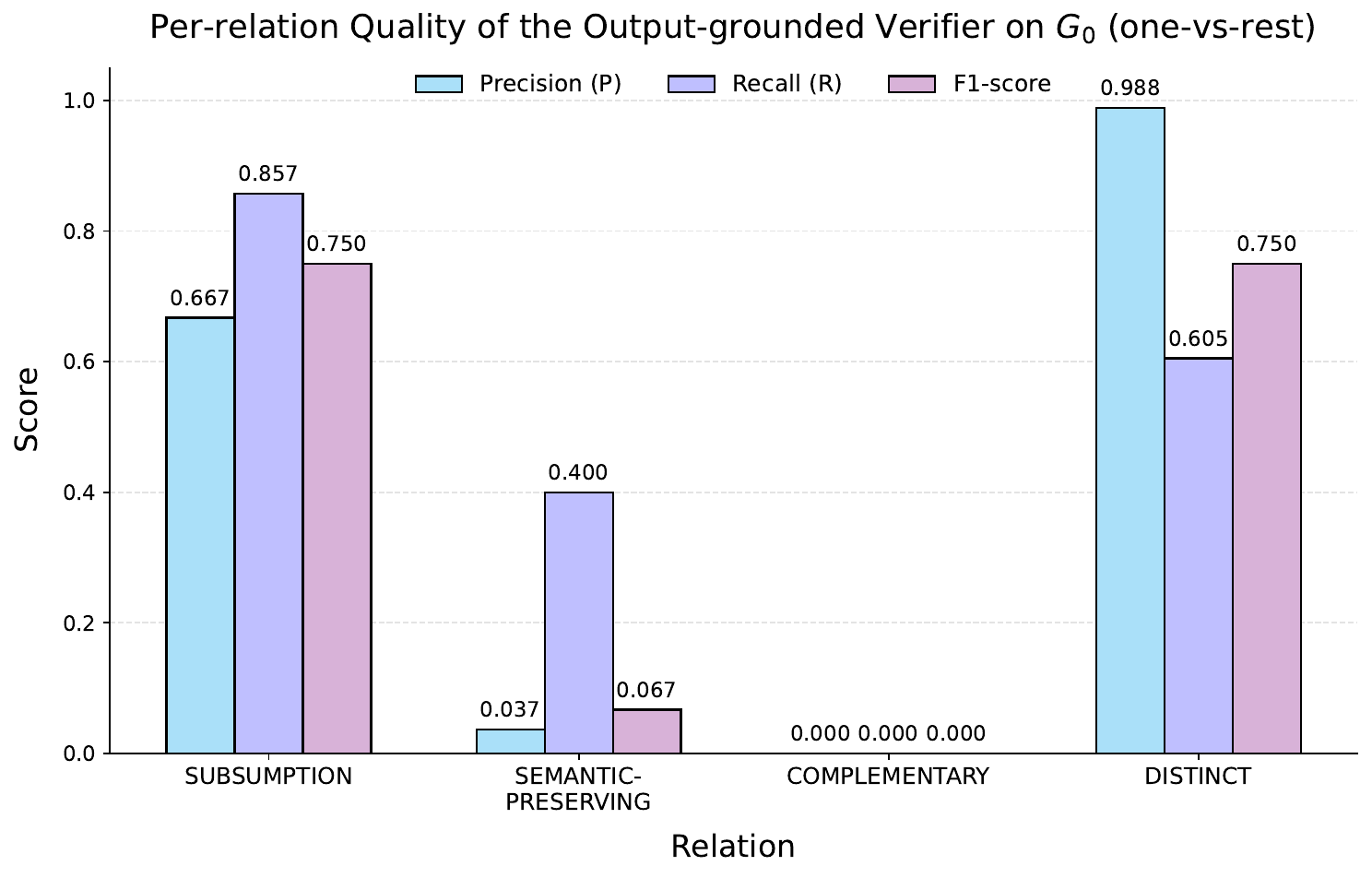}
    \caption{Per-relation precision, recall, and F1 of the output-grounded verifier on $G_0$ (one-vs-rest). \textsc{EXACT} and \textsc{PERCEPTUAL} have no support in a clean reference graph and are omitted.}
    \label{fig:verifier-relations}
\end{figure*}

\medskip
\noindent\textbf{Reading the safety table.} On $G_0$ the false-discovery rate on merges is trivially $1.000$ for every rung that merges anything, because the clean graph contains no true duplicates. The main text therefore reports the false-positive rate on distinct pairs and the hard-negative count instead. For the same reason, precision-recall curves for the mergeable class are undefined on $G_0$, and a $0/6$ hard-negative result bounds the false-merge rate only at $39.3\%$ (one-sided 95\%). The exact 95\% CIs for the hard-negative counts in \cref{tab:verifier-safety} are $[0.36, 1.00]$ for $5/6$, $[0.22, 0.96]$ for $4/6$, $[0.00, 0.64]$ for $1/6$, and $[0.00, 0.46]$ for $0/6$.

\medskip
\noindent\textbf{How wrappers affect code-based comparisons.} The corruption harness creates duplicate skills as wrappers that delegate execution to existing skills. In one fully corrupted instance, 64 synthesized skills reduce to just 4 distinct source strings. Among the $2{,}016$ pairs formed by synthesized skills from different originals, $519$ ($25.7\%$) have identical source code despite wrapping different skills. As a result, code-based methods can miss genuine duplicates when their implementations differ, while incorrectly matching unrelated skills whose wrapper code is identical.

Description-based methods can still be evaluated on the corrupted graphs, but the harness copies each original skill's description directly to its wrapper. This makes duplicate detection easier than in real-world libraries, where duplicated skills may have different descriptions. The output-grounded verifier avoids this particular problem because a wrapper produces the same output as the skill it delegates to.

\medskip
\noindent\textbf{Comparing methods against reference labels.} The parameter-sweep procedure described in \cref{sec:app_real} provides labels for all 182 Corpus-A pairs, enabling precision-recall evaluation on real data. This is not possible on $G_0$, where no pairs are mergeable, or on $G_\rho$, where the wrapper representation makes the source methods unsuitable for comparison. We therefore compare the methods at matched operating points and release the labels with the benchmark. The current evaluation harness supports only built-in skills, so testing third-party implementations requires an extension. Finally, results for the output-grounded verifier should be interpreted as a comparison with exhaustive parameter sweeps, not as independent validation, since both evaluate the same skill outputs.

\subsection{Verifier Self-Validation}
\label{sec:app_selfval}

\noindent\textbf{Answer-key repair.} The corruption log records the relations a defect injects but not the incidental relations cloning creates. Audited over all 300 instances, $1{,}546$ incidental relations were unrecorded against $12{,}790$ recorded ($\approx$12\%), touching 280 of 300 instances. Because unlabeled pairs default to \textsc{DISTINCT}, scoring against the raw key would have penalized the verifier on roughly $1{,}500$ pairs it labels correctly. We repair by soundness rather than convenience. Relations inherited through \textsc{EXACT} clones transfer exactly as the outputs are identical. Because \textsc{PERCEPTUAL} clone relationships depend on the chosen thresholds, they are not always stable. We therefore exclude these pairs rather than assign them potentially unreliable labels (740 of $14{,}506$ labeled pairs across the full grid). The repaired key was recomputed independently from scratch with identical results.

\begin{table}[t]
\centering
\caption{Merge safety by corruption rate $\rho$ on the full $G_\rho$ replay ($n{=}300$ instances). False merges are zero at every rate.}
\label{tab:selfval_rho}
\begin{tabular}{ccccc}
\toprule
$\rho$ & pairs & mergeable & merge recall & false merges \\
\midrule
0.1 & 838 & 60  & 0.433 & 0 \\
0.2 & 950 & 120 & 0.450 & 0 \\
0.3 & 1{,}070 & 180 & 0.544 & 0 \\
0.4 & 1{,}196 & 240 & 0.550 & 0 \\
0.5 & 1{,}300 & 300 & 0.527 & 0 \\
0.6 & 1{,}440 & 370 & 0.535 & 0 \\
0.7 & 1{,}550 & 420 & 0.529 & 0 \\
0.8 & 1{,}652 & 480 & 0.496 & 0 \\
0.9 & 1{,}784 & 540 & 0.504 & 0 \\
1.0 & 1{,}878 & 600 & 0.497 & 0 \\
\bottomrule
\end{tabular}
\end{table}

\medskip
\noindent\textbf{Per-rate safety.} \cref{tab:selfval_rho} shows that the zero-false-merge property holds at all ten corruption rates separately, with no monotone trend in recall. The instance-clustered bootstrap interval for overall merge recall is $[0.497, 0.528]$, consistent with the exact interval in the main text.

\medskip
\noindent\textbf{Results across evaluation sizes.} We first evaluated 34 instances, with four instances per rate, covering $1{,}639$ pairs, of which 436 were mergeable. Merge recall was $0.493$ [$0.445, 0.541$] at a precision of $1.000$. On the full evaluation of 300 instances and $13{,}658$ pairs, recall increased to $0.512$, while the confidence interval narrowed by roughly a factor of three, providing a more precise estimate of merge recall.

The larger evaluation also clarified two observations from the initial results. The 34-instance evaluation produced no \textsc{UNCERTAIN} predictions, whereas the full evaluation produced two among $13{,}658$ pairs. Abstention is therefore extremely rare, but not entirely absent. The initial results also suggested that most missed consolidations involved confusion between \textsc{PERCEPTUAL} and \textsc{SEMANTIC-PRESERVING}. In the full evaluation, 52 \textsc{EXACT} pairs were missed: 32 were classified as \textsc{PERCEPTUAL}, 14 as \textsc{SUBSUMPTION}, and 6 as \textsc{SEMANTIC-PRESERVING}.

For result aggregation, each (instance, pair) combination is counted once, and a successful verdict can correct an earlier error record. Disagreement between two successful verdicts is treated as a determinism bug. No such disagreements occurred.

\subsection{Hard-Negative Expansion and Adjudication}
\label{sec:app_hardneg}

The expanded slice contains 69 engineered hard-negative pairs spanning resampling kernels, border modes, and quantization variants. The verifier makes zero false merges (exact 95\% CI $[0.000, 0.052]$, one-sided 95\% upper bound $4.25\%$), and pairs checked independently by more than one shard agree in every case. On the six original designed pairs, the verifier assigns \textsc{DISTINCT} to one and \textsc{SEMANTIC-PRESERVING} to the other five (the three resize pairs, the padding pair, and the posterize/quantize pair), reproduced identically across two independent runs. Since \textsc{SEMANTIC-PRESERVING} does not permit a merge, the safety claim is unaffected, but we state the labels so the claim cannot be over-read.

\noindent Of the 69 pairs, 26 receive labels that rule out merging but do not classify the skills as \textsc{DISTINCT}. Since it is unclear whether these labels or the intended \textsc{DISTINCT} labels better reflect human judgment, we provide a blinded adjudication package for independent review.

The package includes all 26 pairs for evaluation by three annotators, along with 723 rendered probe images. Pair order and left-right placement are randomized separately for each annotator using recorded seeds. The package also includes the verifier's original labels, difference maps with their amplification factors, a \textsc{cannot-tell} option, and code to calculate Fleiss' $\kappa$ and both bounds. We leave this decision to human annotators rather than use an automated judge, since repeated evaluations by the same model would not provide independent judgments and could overstate agreement. Blinding is also important because anyone familiar with the reported results could otherwise be influenced by knowing which pairs are disputed.

We report the false-merge bound under both interpretations. Using the intended labels, the bound is $4.25\%$ ($0/69$). If human adjudicators agree with the verifier on all 26 disputed pairs, it becomes $6.7\%$ ($0/43$). In both cases, no false merges are observed. The disputed pairs are non-mergeable under either interpretation, so the conclusion remains unchanged.

\subsection{Real-Library Studies}
\label{sec:app_real}

\noindent\textbf{Provenance of the copied package.} Both filters (\texttt{aden} and \texttt{perpetua}) were flagged in an initial smoke test using a single random probe. The same two filters were identified again in the subsequent evaluation using 177 probes. The alignment file is generated automatically rather than written by hand. Since neither filter takes parameters, there are no alignment choices that could have been influenced by the observed outputs. The corpus header also records the order in which the filters were authored. For \texttt{perpetua}, the measured LPIPS distance is $0.2464$ and the DINO p90 distance is $0.0008$. For \texttt{aden}, the corresponding values are $0.0472$ and $0.0250$, respectively.

\medskip
\noindent\textbf{Corpus A composition.} Corpus A holds 93 implementations across 20 operation families, giving 182 within-family pairs, with thresholds carried over from \cref{tab:thresholds} and not recalibrated. Of the 17 pairs that published documentation tables group as directly equivalent, execution certifies 6 \textsc{EXACT} (hash-identical on every matched probe, such as horizontal/vertical flips and posterize across Kornia and torchvision), 4 \textsc{PERCEPTUAL} (such as Gaussian blur and grayscale), 3 \textsc{COMPLEMENTARY}, and 4 \textsc{DISTINCT} (such as CLAHE and histogram-equalization variants with worst-case $L_\infty$ up to $0.94$). Nineteen pairs allow an alternative defensible parameter alignment, and a divergence visible under only one alignment variant is reported as an alignment artifact rather than a behavioral finding.

\medskip
\noindent\textbf{Independent verification using parameter sweeps.} Corpus A has no predefined ground-truth labels, so the verifier's own predictions cannot serve as an independent reference. We instead compare skill outputs across dense parameter sweeps, using tolerances based on numerical precision rather than observed agreement rates ($\epsilon = 1.5/255$, accounting for one 8-bit quantization step and floating-point round-trip error, and $\delta = 0.10$). This procedure classifies all 182 pairs without ambiguity at the pixel-comparison stage: 75 are labeled \textsc{EQUIVALENT} and 107 \textsc{DISTINCT}.

The results also pass the available consistency checks. All 93 self-pairs in Corpus A are classified as \textsc{EQUIVALENT}. For $G_0$, all 100 self-pairs are classified as \textsc{EQUIVALENT}, and all six engineered hard negatives are classified as \textsc{DISTINCT}. The hard-negative check cannot be applied to Corpus A because those pairs are not part of its skill identifiers; we report this limitation explicitly. We use these labels as an independent reference when evaluating text- and code-based baselines. Comparisons with the output-grounded verifier serve a different purpose: they measure agreement between exhaustive parameter-sweep comparisons and the verifier's approximate decisions. We distinguish these two forms of evaluation throughout the paper.

\section{Curation-Agent Supplementary Results}
\label{sec:app_agents}

\cref{tab:app_outcomes} breaks the per-episode action budget into applied, rejected, blocked, and invalid outcomes, and \cref{tab:app_perrho} reports mean F1 at every corruption rate for all nine curators.

\begin{table}[t]
\centering
\caption{Mean per-episode action outcomes. Applied actions changed the library, while rejected, blocked, and invalid actions were discarded by VisCurate's validation checks. GPT-5.5 as in \cref{tab:model_evaluation}.}
\label{tab:app_outcomes}
\resizebox{\columnwidth}{!}{%
\begin{tabular}{l c c c c}
\toprule
\textbf{Model} & {\textbf{Applied}} & {\textbf{Rejected}} & {\textbf{Blocked}} & {\textbf{Invalid}} \\
\midrule
Llama 3.2 1B & 0.12  & 7.32   & 8.83 & 135.84 \\
Llama 3.2 3B & 1.05  & 4.14   & 0.00 & 194.81 \\
Llama 3.1 8B & 4.08  & 7.34   & 0.00 & 188.21 \\
Qwen 3.5 2B  & 0.60  & 3.41   & 0.05 & 124.25 \\
Qwen 3.5 4B  & 11.48 & 46.66  & 0.00 & 127.15 \\
Qwen 3.5 9B  & 3.92  & 134.15 & 0.00 & 33.44  \\
Qwen 3.5 27B & 37.90 & 69.96  & 0.00 & 1.56   \\
Gemma 4 12B  & 13.15 & 158.82 & 0.00 & 0.21   \\
GPT-5.5& 41.42 & 24.41  & 0.00 & 0.01   \\
\bottomrule
\end{tabular}
}
\end{table}

\begin{table*}[t]
\centering
\caption{Mean F1 by corruption rate $\rho$ ($n{=}30$ episodes per cell). Difficulty is not monotone in $\rho$, and the direction is model-dependent.}
\label{tab:app_perrho}
\begin{tabular}{lcccccccccc}
\toprule
\textbf{Model} & $\rho{=}0.1$ & $0.2$ & $0.3$ & $0.4$ & $0.5$ & $0.6$ & $0.7$ & $0.8$ & $0.9$ & $1.0$ \\
\midrule
GPT-5.5      & 0.420 & 0.521 & 0.579 & 0.596 & 0.621 & 0.641 & 0.624 & 0.628 & 0.625 & 0.612 \\
Qwen 3.5 27B & 0.457 & 0.476 & 0.522 & 0.559 & 0.523 & 0.512 & 0.524 & 0.519 & 0.484 & 0.474 \\
Gemma 4 12B  & 0.378 & 0.384 & 0.356 & 0.353 & 0.321 & 0.252 & 0.231 & 0.209 & 0.210 & 0.206 \\
Qwen 3.5 4B  & 0.216 & 0.226 & 0.185 & 0.160 & 0.120 & 0.068 & 0.048 & 0.033 & 0.017 & 0.018 \\
Qwen 3.5 9B  & 0.332 & 0.229 & 0.176 & 0.098 & 0.055 & 0.043 & 0.025 & 0.007 & 0.027 & 0.018 \\
Llama 3.1 8B & 0.205 & 0.186 & 0.124 & 0.118 & 0.079 & 0.067 & 0.049 & 0.053 & 0.038 & 0.028 \\
Llama 3.2 3B & 0.067 & 0.032 & 0.021 & 0.016 & 0.024 & 0.021 & 0.020 & 0.014 & 0.014 & 0.007 \\
Qwen 3.5 2B  & 0.049 & 0.039 & 0.020 & 0.006 & 0.005 & 0.005 & 0.002 & 0.003 & 0.002 & 0.001 \\
Llama 3.2 1B & 0.000 & 0.000 & 0.000 & 0.000 & 0.000 & 0.000 & 0.000 & 0.000 & 0.000 & 0.000 \\
\bottomrule
\end{tabular}%
\end{table*}

\medskip
\noindent\textbf{Results on the hardest instances.} GPT-5.6-luna was evaluated on 41 of the most difficult instances, covering all cases with $\rho{=}1.0$ and the hardest cases with $\rho{=}0.9$. It achieved a mean F1 of $0.6201$ [$0.5946, 0.6471$], with precision $0.800$, recall $0.517$, an Intrinsic Score of $0.506$, and an average action cost of $148.7$. On the same instances, Qwen 3.5 27B achieved an F1 of approximately $0.48$, while all models with 9B parameters or fewer scored below $0.035$. We report these results separately from the main table because they cover only 41 difficult instances, whereas the main results use the full 300-instance evaluation.

\begin{figure*}[t]
\centering
\begin{tikzpicture}
\begin{axis}[
    ybar,
    ymin=0, ymax=1.1,
    xmin=0.4, xmax=9.6,
    enlarge x limits=false,
    ylabel={\% of Runs Ended},
    title={\textbf{Ended Rate by Model Family}},
    width=0.85\textwidth,
    height=6.5cm,
    bar width=0.9cm,
    clip=false,
    ymajorgrids=true,
    grid style={dashed, gray!30},
    xtick={1,2,3,4,5,6,7,8,9},
    xticklabels={
        Gemma 4 12B,
        Llama 3.2 3B,
        Llama 3.1 8B,
        Llama 3.2 1B,
        OpenAI GPT 5.5,
        Qwen 3.5 4B,
        Qwen 3.5 9B,
        Qwen 3.5 2B,
        Qwen 3.5 27B
    },
    xticklabel style={rotate=45, anchor=east, font=\small},
    axis x line=bottom,
    axis y line=left,
    nodes near coords,
    nodes near coords align={vertical},
    every node near coord/.append style={font=\footnotesize},
    point meta=y,
    /pgf/number format/.cd,
    fixed,
    fixed zerofill,
    precision=3
]
\addplot[fill=cyan!30, draw=black, bar shift=0pt] coordinates {
    (1, 0.197)
};
\addplot[fill=blue!25, draw=black, bar shift=0pt] coordinates {
    (2, 0.000)
    (3, 0.000)
    (4, 0.243)
};
\addplot[fill=violet!30, draw=black, bar shift=0pt] coordinates {
    (5, 0.965)
};
\addplot[fill=green!30, draw=black, bar shift=0pt] coordinates {
    (6, 0.157)
    (7, 0.207)
    (8, 0.403)
    (9, 0.743)
};
\end{axis}
\end{tikzpicture}
\caption{Percentage of runs that the models have ended using the \textsc{END} action.}
\label{fig:ended-rate}
\end{figure*}
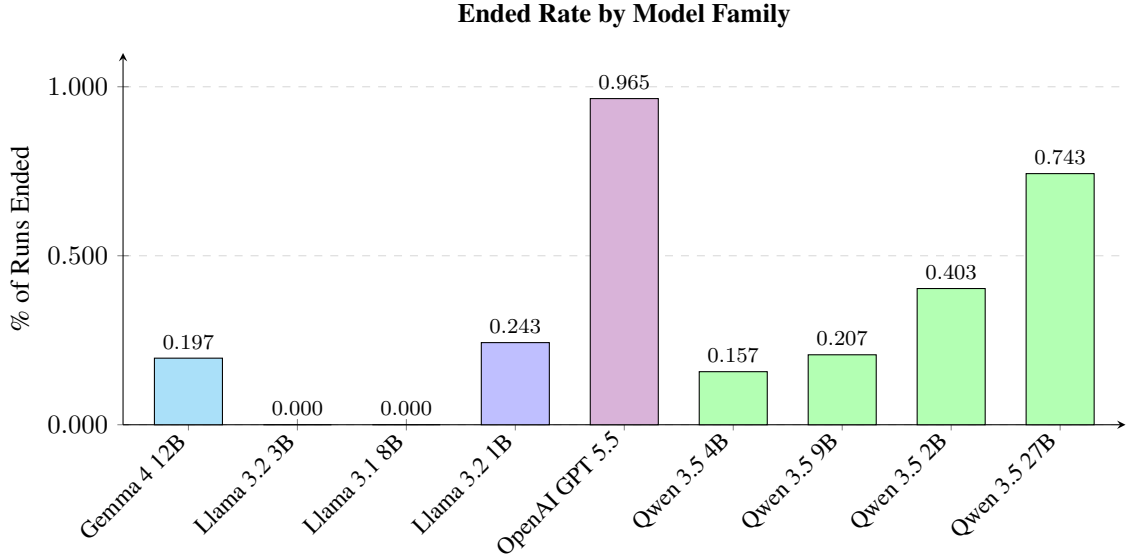

\section{Cost and Scalability Details}
\label{sec:app_cost}

\begin{figure*}
    \centering
    \includegraphics[width=0.7\linewidth]{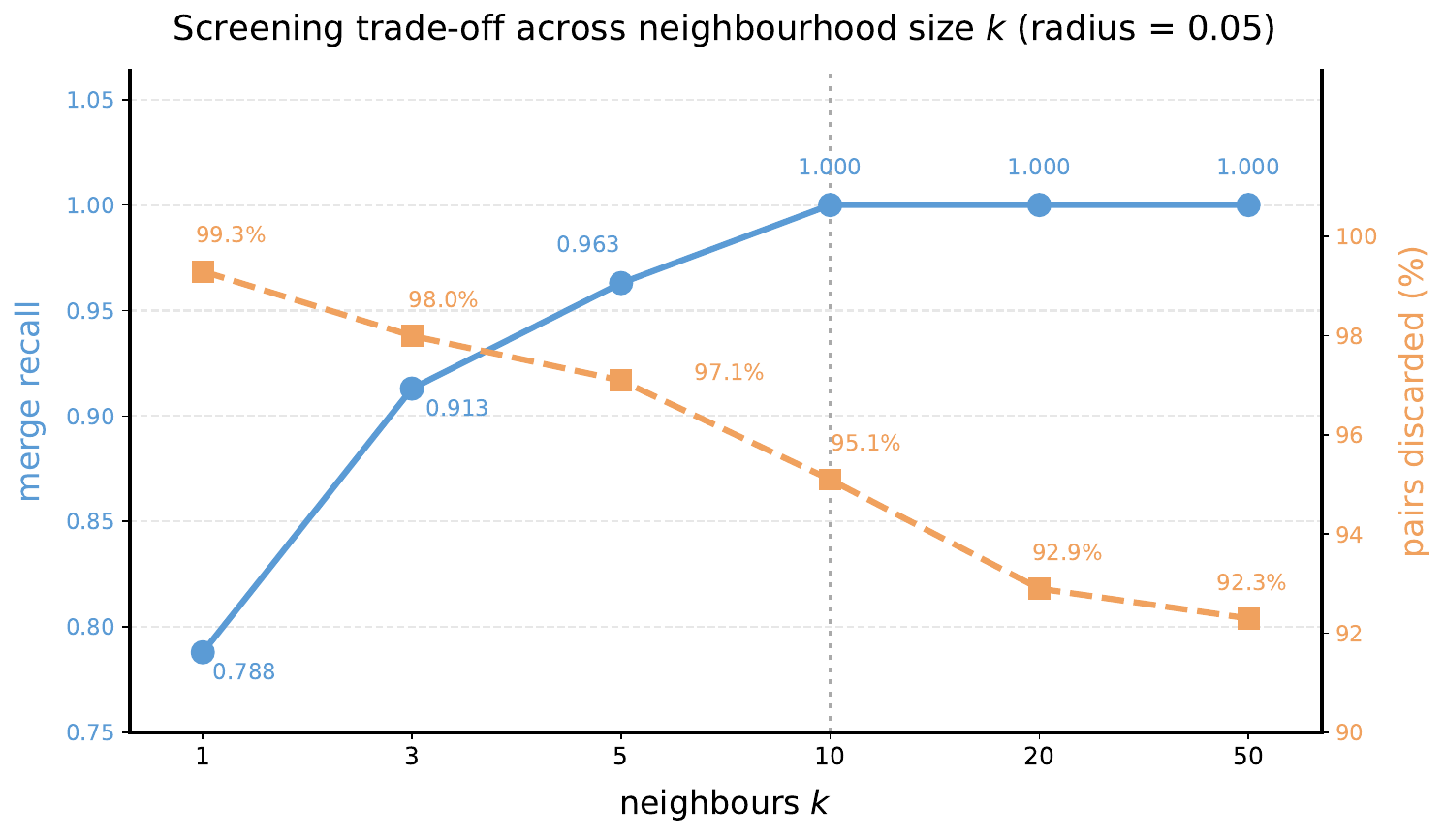}
    \caption{Screening trade-off across neighbor count $k$ at radius $0.05$, over ten corrupted instances (80 mergeable positives, $76{,}031$ pairs). Merge recall saturates at $k{=}10$ while $95.1\%$ of pairs are still discarded, and the curves are insensitive to radius from $0.05$ to $2.0$.}
    \label{fig:app_screen}
\end{figure*}

\begin{figure*}
    \centering
    \includegraphics[width=\linewidth]{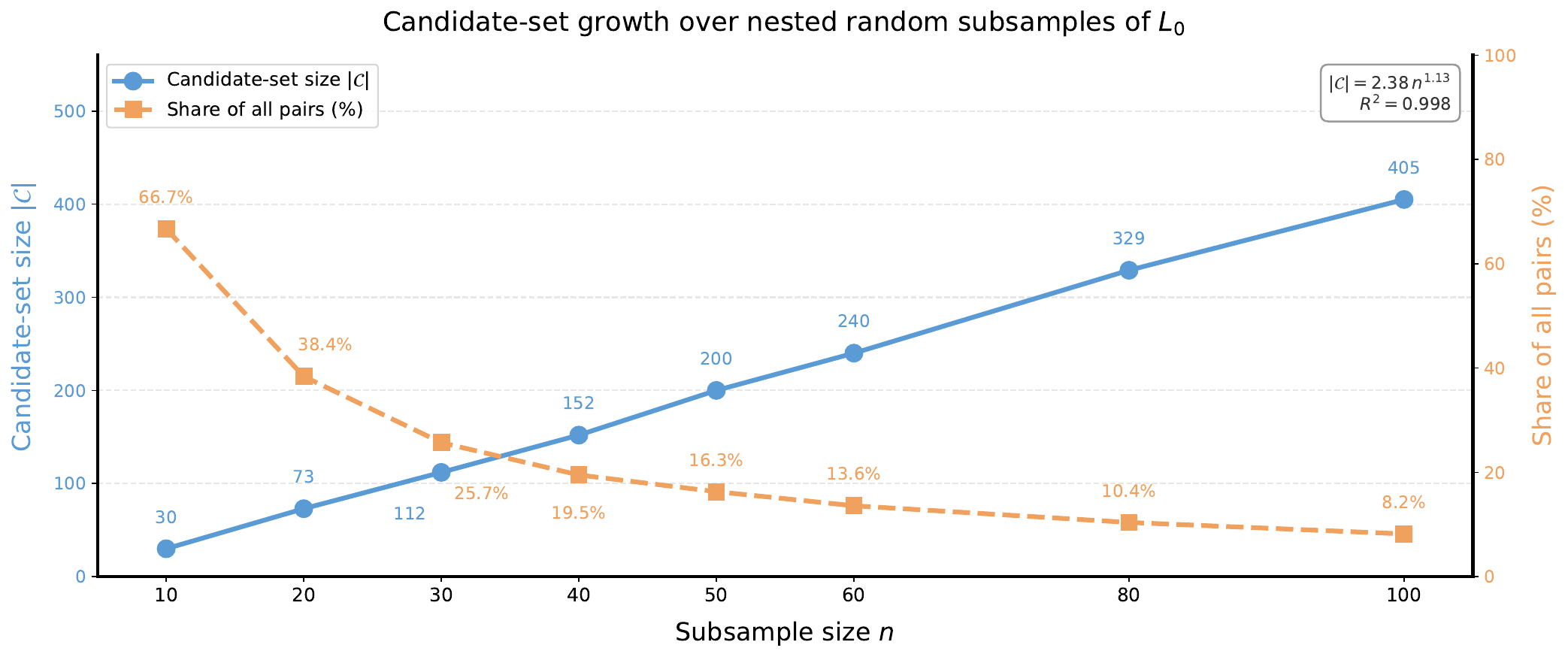}
    \caption{Candidate-set growth over nested random subsamples of $L_0$. The log-log fit $|C| = 2.38\,n^{1.13}$ ($R^2 = 0.998$) is close to linear, and the share of all pairs that screening proposes falls from $66.7\%$ at $n{=}10$ to $8.2\%$ at $n{=}100$.}
    \label{fig:app_growth}
\end{figure*}

\begin{figure*}
    \centering
    \includegraphics[width=0.9\linewidth]{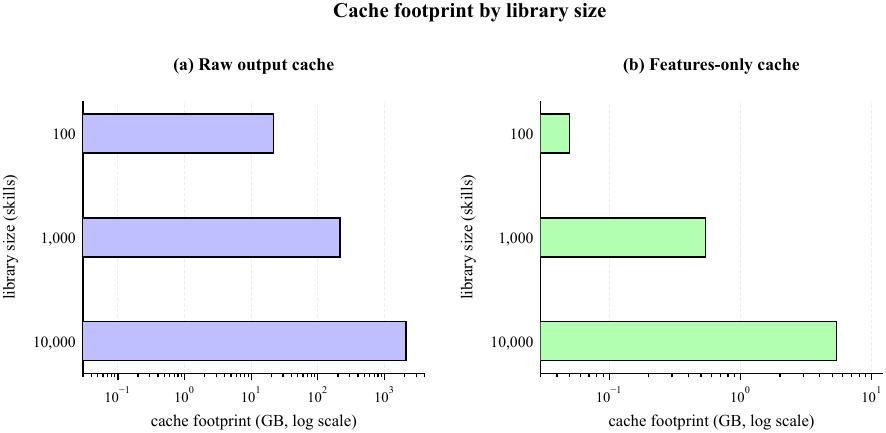}
    \caption{Cache footprint by library size for (a) raw output sets and (b) backend features only. The measured unit is $214.8$\,MB per skill for raw outputs over the 177-probe battery (median of 8 builtin skills), against an analytic features-only cache, a $395\times$ saving at $10^4$ skills.}
    \label{fig:libraryFootprint}
\end{figure*}

\cref{fig:app_screen} reports the screening sweep behind the $k{=}10$ operating point, \cref{fig:app_growth} the near-linear growth of the candidate set, \cref{tab:app_stages} which cascade stage decides each pair, and \cref{fig:libraryFootprint} the cache footprint that motivates caching features instead of raw outputs.



\begin{table}[t]
\centering
\caption{Which cascade stage decides each pair ($12{,}745$ decisions). A majority is settled before any learned backend runs.}
\label{tab:app_stages}
\begin{tabular}{lcc}
\toprule
\textbf{Deciding stage} & $n$ & \textbf{share} \\
\midrule
residual (distinct by elimination) & 6{,}020 & 47.2\% \\
semantic       & 2{,}672 & 21.0\% \\
subsumption    & 2{,}032 & 15.9\% \\
complementary  & 938  & 7.4\% \\
exact          & 918  & 7.2\% \\
perceptual     & 110  & 0.9\% \\
abstention band & 55  & 0.4\% \\
\bottomrule
\end{tabular}
\end{table}

\begin{table}[t]
\centering
\caption{Token cost per correct repair on the 41 telemetry-instrumented episodes shared by both hosted models. The denominator is correct repairs, not actions applied.}
\label{tab:app_tokens}
\resizebox{\columnwidth}{!}{%
\begin{tabular}{lccccc}
\toprule
\textbf{Model} & \textbf{actions} & \textbf{applied} & \textbf{correct} & \textbf{tokens (in/out)} & \textbf{tok./repair} \\
\midrule
GPT-5.5      & 4{,}233 & 2{,}537 & 2{,}200 & 18.3M / 1.96M & \textbf{9{,}202} \\
GPT-5.6-luna & 6{,}122 & 2{,}994 & 2{,}338 & 22.5M / 2.39M & 10{,}667 \\
\bottomrule
\end{tabular}%
}
\end{table}

Three accounting rules are followed for \cref{tab:app_tokens}. First, the numerator and denominator must come from the same episodes. Dividing all-grid repairs by instrumented-episode tokens understates the true cost by up to $5\times$, which we verified by making that mistake and correcting it. Second, cost per repair is undefined, not large, for a model with zero correct repairs. Third, telemetry was instrumented after the eight self-hosted cells of the main table ran, so their token costs are not reported. Re-running $8\times300$ episodes solely to backfill a cost column would change the cells' provenance, and we prefer the stated gap. Verifier-side telemetry covers $262{,}833$ events across 156 files, with a signature-cache hit rate of $78.2\%$ ($184{,}824/236{,}305$) and a per-pair decision latency of median $401$\,ms and p95 $64.5$\,s. Eviction from the byte-bounded cache never changes a verdict, only wall-clock, so memory can be traded for time freely.

\section{Extension Studies}
\label{sec:app_extensions}

Our executable core is deliberately restricted to deterministic classical operations, while modern agent ecosystems also lean on stochastic and neural tools~\cite{ravi2024sam2,liu2024grounding,yang2024depth,rombach2022high}. The three studies below extend the protocol to such skills, measure certificate stability on out-of-distribution probes, and ablate the probe battery.

\subsection{Neural and Stochastic Skills}
\label{sec:app_neural}

\noindent\textbf{Evaluation procedure.} (1) Comparison is conditional per probe, never over pooled distributions, since a model applying $\mathrm{blur}$ to $x_1$ and $\mathrm{sharpen}$ to $x_2$ and one doing the reverse have identical pooled output distributions while behaving differently, so pooling would manufacture a false merge by construction. (2) Equality is tested by a permutation null rather than a naive bootstrap, because the unbiased MMD estimator~\cite{gretton2012kernel} can be negative and degenerates under the null, exactly where equivalence decisions live. (3) Depth outputs get per-image affine (scale and shift) alignment so output conventions are not measured as behavior. (4) We do not claim the deterministic $0.99$ precision calibration transfers, and report counts with exact bounds. The errors of this path are the errors of a hypothesis test rather than the tolerance-bounded verdicts of the deterministic path.

\medskip
\noindent\textbf{Setup and validity checks.} We compare six neural skills over 40 probes, with 16 stochastic draws per probe, 2{,}000 permutations, and $\alpha=0.05$. The skills are SAM~2.1 hiera large and small for promptable segmentation, Depth Anything V2 large and small for monocular depth, and two purpose-built hard negatives, a disparity variant and a semantic-mask variant. Both validity checks pass. All 6/6 self-pairs are not distinguishable from themselves (the noise floor), and both 2/2 hard negatives are distinguished.

\begin{table}[t]
\centering
\caption{Distributional comparison of neural skills (permutation-null MMD, per probe). ``Applicable'' restricts to object-bearing images, the domain where segmentation is defined.}
\label{tab:app_neural}
\resizebox{\columnwidth}{!}{%
\begin{tabular}{llll}
\toprule
\textbf{Comparison} & \textbf{Domain} & \textbf{Verdict} & \textbf{min $p$} \\
\midrule
DAv2-Large vs DAv2-Small & full & distinguishable & 0.0005 \\
SAM2-Large vs SAM2-Small & full & distinguishable & 0.0005 \\
SAM2-Large vs SAM2-Small & applicable & \textbf{not disting.} & 0.0355 \\
depth vs disparity (hard neg.) & full & distinguishable & 0.0005 \\
instance vs semantic (hard neg.) & applicable & distinguishable & 0.0010 \\
\bottomrule
\end{tabular}%
}
\end{table}

\medskip
\noindent\textbf{Findings.} The two depth models are distinguishable on all 40 probes (\cref{tab:app_neural}). SAM~2 large and small separate on synthetic patterns (13 of 38 probes reject) but not on object-bearing images (0 of 10), so two capacities of the same architecture are not distinguishable at this power exactly where segmentation is defined. A non-rejection is absence of evidence, never equivalence. The instance-vs-semantic negative cannot show up on object-free inputs, where masks collapse to one region and make ``best mask'' and ``union of masks'' identical by construction, and with 137 of 177 probes synthetic, the default battery is not a valid domain for segmentation skills. Both domains are reported, and we disclose that the restriction to object-bearing images was chosen after seeing the full-battery result.  The disparity variant was distinguishable from \emph{itself} because $1/\mathrm{clip}(d, 10^{-3})$ explodes near zero depth, letting one-LSB jitter dominate the normalized map. It was fixed with a 5th-percentile floor, the unstable run is archived, and we corrected a tally of the validity checks that had silently dropped the depth-vs-disparity row, reporting 1/1 instead of 2/2.

\subsection{Testing Certified Merges on Unseen Inputs}
\label{sec:app_ood}

We evaluate the verifier on $B_{\mathrm{ood}}$, a set of 31 probes covering six classes absent from the original distribution: extreme aspect ratios, HDR, medical, satellite, saturated, and tiny images. The thresholds remain unchanged. We measure how often pairs labeled \textsc{EXACT} or \textsc{PERCEPTUAL} on $B$ receive a different label on $B_{\mathrm{ood}}$.

Of the 74 merges accepted on the real libraries (\cref{subsec:real}), only one changes its label, from \textsc{PERCEPTUAL} to \textsc{SEMANTIC-PRESERVING}. This gives a flip rate of $0.0135$, with an exact 95\% confidence interval of $[0.0003, 0.0730]$ and a one-sided 95\% upper bound of $6.25\%$.

Although the observed flip rate is low, the result shows that some merges accepted on the original probes may not hold for unseen inputs. We focus on these cases because agreement on additional inputs alone does not establish equivalence. We do not report reverse flips, since a previously rejected pair appearing mergeable on new inputs is not sufficient evidence that the pair is equivalent.

\subsection{Probe-Battery Ablation}
\label{sec:app_ablation}

We evaluate 21 configurations on the 24 designed $G_0$ pairs, including six engineered hard negatives. The experiments vary battery size ($|B| \in \{8,16,32,64,128,177\}$), selection strategy, and verifier backend. The subsets are nested, and selection depends only on class, index, or seed, not on the observed results. Accuracy remains similar across battery sizes, with overlapping intervals of roughly $0.33$ to $0.58$ at $n{=}24$. The results show no clear improvement from increasing the battery size.

\begin{table}[t]
\centering
\caption{Hard-negative false merges by battery size and probe-selection strategy. Single-class selection produces a false merge that stratified and random selection never do.}
\label{tab:app_ablation}
\begin{tabular}{lcccccc}
\toprule
\textbf{Strategy} & $|B|{=}8$ & 16 & 32 & 64 & 128 & 177 \\
\midrule
stratified & 0 & 0 & 0 & 0 & 0 & 0 \\
first (single-class) & \textbf{1} & \textbf{1} & \textbf{1} & \textbf{1} & 0 & 0 \\
random & 0 & 0 & 0 & 0 & 0 & 0 \\
\bottomrule
\end{tabular}
\end{table}

Safety varies with how probes are selected (\cref{tab:app_ablation}). Because probes are ordered by class, selecting the first $N$ can introduce a hard-negative false merge that persists through $|B|{=}64$. Neither stratified nor random selection produces this error. This suggests that probe coverage matters more than battery size alone, and supports selecting probes across classes rather than by index.

In the full-size backend ablation, pixel-only comparison achieves similar accuracy on designed relations to the full stack ($0.50$ vs. $0.46$), with no hard-negative false merges in any configuration. This is consistent with the finding that $54.9\%$ of pairs are decided before learned backends are used. We treat the full battery as a reference, not absolute ground truth: disagreements with a smaller battery indicate less evidence, while correctness is assessed against the designed relations.

\section{Auditing Untrusted Repositories}
\label{sec:app_comfyui}

\noindent Auditing collections such as ComfyUI node packs requires executing untrusted code, which is not permitted on shared infrastructure. Such audits require disposable, isolated environments with no network access, host credentials, or writable host mounts.

We recommend stratified sampling of popular and long-tail skills, testing each pack separately, and reporting failures at import, execution, and contract validation. Contract violations count as failures rather than being clamped, and compression is measured through sequential curation because \textsc{PERCEPTUAL} relationships are not necessarily transitive. We leave this audit to suitably isolated environments; the real-library studies in \cref{subsec:real} provide complementary evidence.

\end{document}